\documentclass{article}

\usepackage[preprint]{neurips_2026}

\usepackage[utf8]{inputenc} 
\usepackage[T1]{fontenc} 
\usepackage{hyperref} 
\usepackage{url} 
\usepackage{booktabs} 
\usepackage{amsfonts} 
\usepackage{nicefrac} 
\usepackage{microtype} 
\usepackage{xcolor} 
\usepackage{graphicx}
\usepackage{amsmath}
\usepackage{amssymb}

\title{HarmoGS: Robust 3D Gaussian Splatting in the Wild via Conflict-Aware Gradient Harmonization}

\author{%
Yulei Kang\textsuperscript{1}\thanks{Equal contribution.} \quad
Tianze Zhu\textsuperscript{2}\footnotemark[1] \quad
Jian-Fang Hu\textsuperscript{1,3,4}\thanks{Corresponding author.} \quad
Jianhuang Lai\textsuperscript{1,3,4} \quad
Wei-Shi Zheng\textsuperscript{1,4} \\
\textsuperscript{1} Sun Yat-sen University \quad \textsuperscript{2} Northeastern University\\
\textsuperscript{3} Guangdong Province Key Laboratory of Information Security Technology, China \\
\textsuperscript{4} Key Laboratory of Machine Intelligence and Advanced Computing, Ministry of Education, China\\
}

\begin{document}
  \maketitle
  
\begin{abstract}
In-the-wild 3D Gaussian Splatting remains challenging due to transient distractors and illumination-induced cross-view appearance inconsistencies. Existing methods mainly rely on image-level masking to suppress unreliable supervision, but masking alone cannot fully eliminate residual occlusions or resolve illumination-induced inconsistencies, both of which can introduce conflicting cross-view gradients. These unresolved conflicts may destabilize Gaussian optimization and lead to visible reconstruction artifacts. We propose a conflict-aware 3DGS framework that addresses this problem from both image-space supervision and gradient-level optimization. Semantic Consistency-Guided Masking learns pixel-wise consistency scores to adaptively refine prior masks and suppress unreliable supervision before gradient formation. A dual-view Conflict-Aware Gradient Harmonization strategy further reconciles view-specific gradients by mutually rotating them into an orthogonal configuration, reducing negative directional interference across views. We also introduce conflict-aware densification and pruning to stabilize Gaussian growth and remove persistently conflicting primitives. Extensive experiments on standard in-the-wild benchmarks demonstrate that our method achieves state-of-the-art rendering quality under complex transient distractors and cross-view inconsistencies.
\end{abstract}

\section{Introduction}
Reconstructing a reliable 3D representation from a collection of images is a fundamental problem in computer vision, with broad applications in VR/AR, film production, and game development. Recent advances, including Neural Radiance Fields (NeRF) \cite{mildenhall2021nerf} and 3D Gaussian Splatting (3DGS) \cite{kerbl20233d}, have achieved remarkable success in photorealistic scene modeling. However, these methods typically assume that training images are captured under consistent illumination and contain stable scene content, which rarely holds in unconstrained real-world environments.

In in-the-wild scenes, captured images are often affected by transient distractors and appearance variations, such as moving pedestrians, vehicles, and illumination changes. Existing methods mainly address this issue by suppressing unreliable image-level supervision. Some approaches \cite{zhang2024gaussian, ren2024nerf, ungermann2024robust, sabour2023robustnerf} learn to down-weight transient regions in a self-supervised manner, but they often struggle with large or semantically complex occlusions. Recent methods \cite{tang2025dronesplat, li2025robust} further use pretrained segmentation models to generate object-level mask candidates, and then identify dynamic distractors with hand-crafted criteria, such as reconstruction residual thresholds or temporal tracking. However, these rule-based pipelines remain sensitive to segmentation quality and heuristic design, limiting their robustness in complex real-world scenes. To address these limitations, we introduce Semantic Consistency-Guided Masking, which learns a pixel-wise consistency score during training and uses it to adaptively refine a precomputed SAM-based \cite{kirillov2023segment} binary prior mask. This design combines adaptive reliability estimation with object-level semantic priors, making the mask more robust to complex transient occlusions and initial segmentation errors. By suppressing unreliable image evidence before gradient formation, it reduces noisy gradients in Gaussian optimization.

\begin{figure}[t]
\centerline{\includegraphics[width=1.0\textwidth]{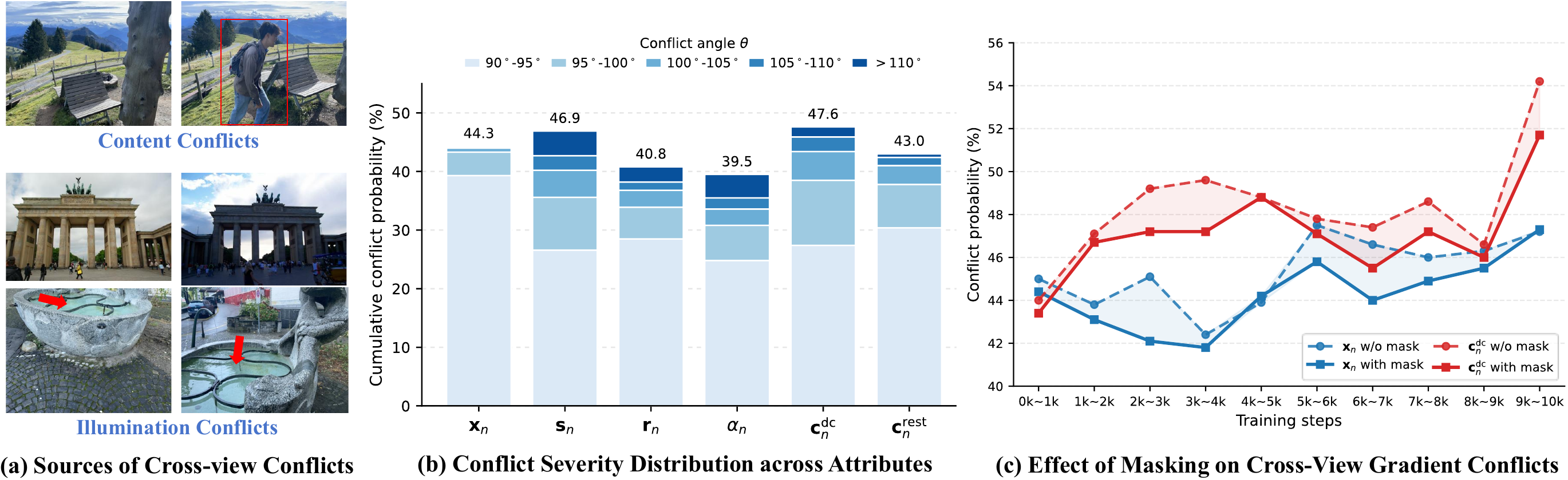}}
\caption{
(a) In-the-wild image collections often contain content conflicts caused by transient objects and illumination conflicts across views, which introduce unreliable supervision.
(b) When two views are jointly used to optimize a shared Gaussian representation, their induced gradients often conflict across different Gaussian attributes.
(c) Applying masks to unreliable image regions only marginally reduces the conflict probability during training, while a substantial level of cross-view conflict still persists.
Statistics are reported on the NeRF On-the-go dataset~\cite{ren2024nerf}.
}
\label{fig:intro}
\vspace{-3mm}
\end{figure}

Although this masking strategy alleviates noisy supervision, image-space masking alone cannot fully resolve optimization conflicts in in-the-wild 3DGS. As shown in Figure~\ref{fig:intro}(b) and (c), cross-view gradient conflicts are common during Gaussian optimization and remain significant even after unreliable regions are masked. This is because masks may fail to fully exclude complex occlusions, while illumination inconsistencies introduce view-dependent supervision that cannot be handled by masking. Consequently, inconsistent cross-view supervision can still produce incompatible Gaussian updates, potentially destabilizing optimization and leading to visible reconstruction artifacts. Motivated by this observation, we introduce a dual-view 3DGS training framework with Conflict-Aware Gradient Harmonization. At each iteration, two training views are randomly sampled to compute view-specific gradients for Gaussian attributes. We then harmonize the two gradients by mutually rotating their directions into an orthogonal configuration. This explicitly eliminates negative directional interference between the two views while largely preserving their original descent directions, yielding a more compatible update for Gaussian optimization.

Building on this framework, we further introduce conflict-aware densification and pruning. Our densification strategy aggregates harmonization-aware view-space gradients from both views to guide Gaussian growth with more reliable two-view signals. The pruning strategy tracks per-Gaussian cross-view conflicts and periodically decays the opacity of highly conflicting primitives, allowing persistently unstable Gaussians to be naturally removed by the standard pruning process.

Our contributions are summarized as follows:
1) We propose a conflict-aware gradient harmonization framework for robust in-the-wild 3D Gaussian Splatting. By explicitly reconciling cross-view gradient conflicts, our method reduces incompatible Gaussian updates caused by transient distractors and illumination-induced inconsistencies.
2) We introduce Semantic Consistency-Guided Masking, which adaptively refines the prior mask and effectively suppresses transient distractors. We further develop conflict-aware densification and pruning strategies to stabilize Gaussian optimization under dual-view training, leading to more reliable Gaussian growth and pruning.
3) Extensive Experiments on standard in-the-wild benchmarks demonstrate that our method effectively handles complex transient distractors and cross-view inconsistencies, achieving state-of-the-art rendering quality and reconstruction fidelity.

\section{Related Works}
\textbf{Novel view synthesis.} Synthesizing photorealistic images from novel viewpoints remains a fundamental yet challenging problem in machine learning and graphics. A seminal work, Neural Radiance Fields (NeRF) \cite{mildenhall2021nerf}, represents scenes implicitly by parameterizing color and density with multilayer perceptrons (MLPs), achieving remarkable rendering quality. Building upon this paradigm, subsequent efforts have primarily focused on improving efficiency, either through compact scene representations \cite{chen2022tensorf, sun2022direct, hu2022efficientnerf, muller2022instant, fridovich2022plenoxels, takikawa2021neural, xu2022point} or via network compression techniques \cite{gordon2023quantizing, reiser2021kilonerf}. More recently, 3D Gaussian Splatting (3DGS) \cite{kerbl20233d} has emerged as a powerful explicit representation for novel view synthesis, offering real-time rendering while maintaining high reconstruction quality. This has motivated a series of follow-up studies that improve 3DGS in terms of rendering fidelity, optimization robustness, and scalability to complex scenes \cite{guedon2024sugar, liu2025maskgaussian, lin2024vastgaussian}.

\textbf{In-the-Wild Novel View Synthesis.} Novel view synthesis in wild scenes involves generating views from sparse images affected by lighting variations and transient distractors, posing challenges for stable scene reconstruction. NeRF-W \cite{martin2021nerf} extended NeRF to handle unconstrained internet image collections, learning image-specific appearance embeddings to model lighting variations and using a transient field to address dynamic occlusions. Subsequent works \cite{yang2023cross, chen2024nerf, chen2022hallucinated, ren2024nerf} built on this, enhancing modeling performance and transient object removal. Ha-NeRF \cite{chen2022hallucinated} replaced the 3D transient volume with 2D visibility maps, using CNN-based global appearance encoding to improve appearance modeling. NeRF On-the-go \cite{ren2024nerf} applied uncertainty-guided transient filtering using DINOv2 features, enabling transient removal on video sequences without semantic priors. RobustNeRF \cite{sabour2023robustnerf} treated dynamic interference as outliers in a robust optimization framework, using iteratively reweighted least-squares (IRLS) to down-weight transient pixels. Despite these advances, these methods still suffer from NeRF's inherent issues, such as long training times and slow rendering. 3DGS-based methods \cite{zhang2024gaussian, kulhanek2024wildgaussians, li2025robust} provide an effective solution to these limitations. Wild-GS \cite{xu2024wild} aligns appearance information from reference images to 3D Gaussians via triplane cropping, modeling high-frequency details while maintaining 3DGS's real-time performance. To better handle transient interference, SpotLessSplats \cite{sabour2025spotlesssplats} combines semantic segmentation features with iterative reconstruction residuals to adaptively identify transient regions. HybridGS \cite{lin2025hybridgs} introduces a dual-branch architecture for static and transient elements, improving the distinction between transient objects and static backgrounds. Robust3DGS \cite{ungermann2024robust} leverages the IRLS technique from RobustNeRF \cite{sabour2023robustnerf}, further incorporating image residual clustering for self-supervised transient masking.

\begin{figure}[t]
\centerline{\includegraphics[width=1.0\textwidth]{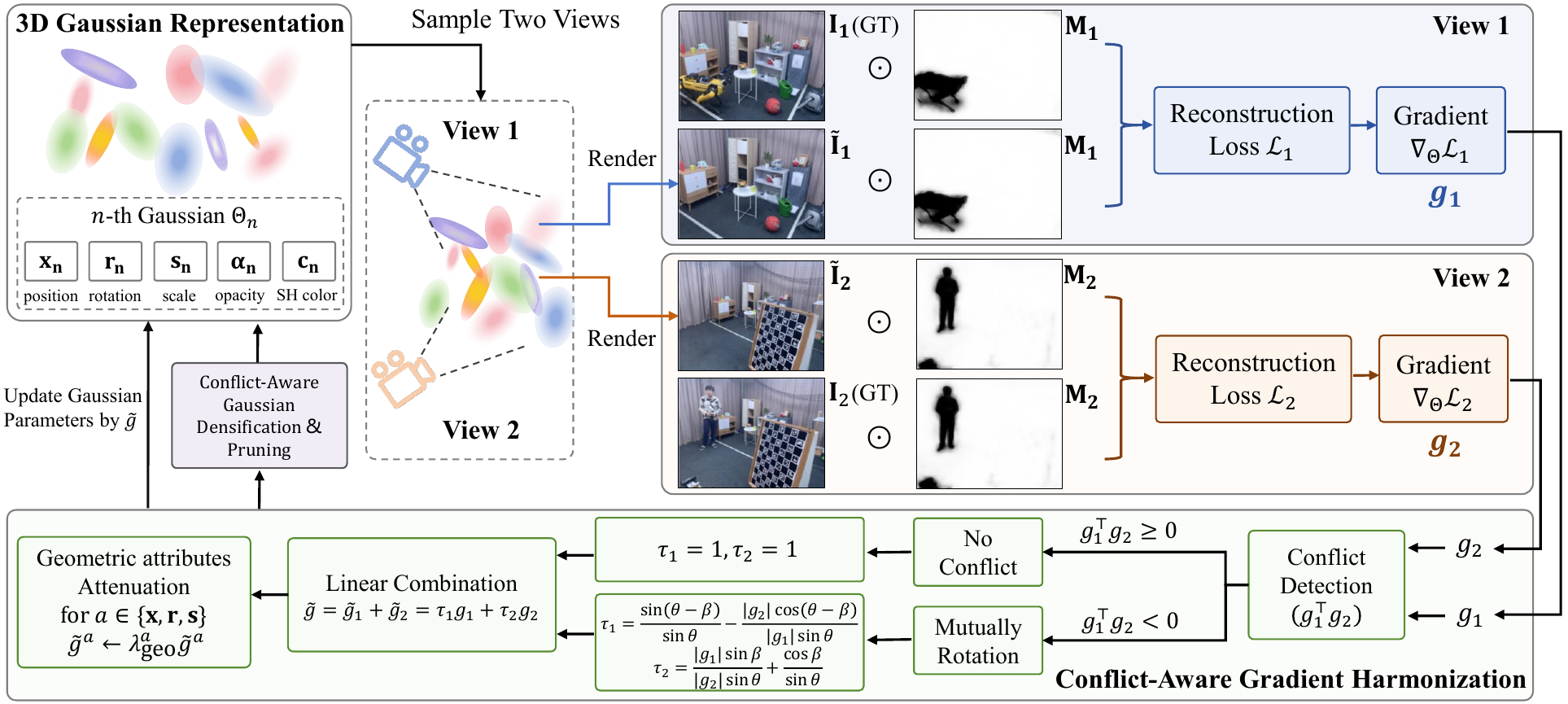}}
\caption{
Overview of the proposed framework. At each training iteration, two views are sampled and rendered from the current 3D Gaussians. Semantic Consistency-Guided Masking generates view-specific masks to suppress transient regions, and the masked reconstruction losses $\mathcal{L}_1$ and $\mathcal{L}_2$ produce the corresponding gradients $g_1$ and $g_2$. When a cross-view conflict is detected, Conflict-Aware Gradient Harmonization reconciles the two gradients by mutually rotating them toward orthogonal directions, which is implemented by computing conflict-aware combination coefficients. The harmonized gradients from the two views are then summed to update the Gaussian parameters.
}
\label{fig:method_primirary}
\vspace{-3mm}
\end{figure}

\section{Method}
\subsection{Preliminaries}
\label{sec:preliminaries}
3D Gaussian Splatting \cite{kerbl20233d} represents a scene with a set of anisotropic Gaussian primitives. Each Gaussian is parameterized by learnable attributes $\Theta_n=\{\mathbf{x}_n,\mathbf{s}_n,\mathbf{r}_n,\alpha_n,\mathbf{c}_n\}$, where $\mathbf{x}_n\in\mathbb{R}^3$ is the center position, $\mathbf{s}_n\in\mathbb{R}^3$ is the scale, $\mathbf{r}_n\in\mathbb{R}^4$ is the rotation quaternion, $\alpha_n\in\mathbb{R}$ is the opacity, and $\mathbf{c}_n$ denotes the spherical-harmonic color coefficients. Given a camera view, Gaussians are projected onto the image plane and rendered by alpha compositing:
\begin{equation}
\mathbf{C} = \sum_{n=1}^{N} \mathbf{c}_n w_n \prod_{m=1}^{n-1}(1-w_m),
\end{equation}
where $w_n$ denotes the per-pixel contribution weight of the $n$-th projected Gaussian, determined by its learnable opacity and its projected 2D Gaussian response at the pixel.

\subsection{Joint Optimization of Cross-View Gaussian Parameters}
Our key insight is that in-the-wild training images often contain cross-view inconsistencies, including illumination conflicts across views and content conflicts caused by transient distractors. When optimizing the model with such inconsistent observations, the gradients derived from different views may point to conflicting update directions, leading to unstable optimization and reconstruction artifacts. To mitigate this, we jointly consider two different training views at each iteration and introduce a gradient harmonization strategy, detailed in Section \ref{sec:gradient_harmonization}, to reconcile their update directions.

Specifically, at each training iteration, we independently sample two views $\mathbf{V}_1$ and $\mathbf{V}_2$ with their corresponding ground-truth images $\mathbf{I}_1$ and $\mathbf{I}_2$. Let $\tilde{\mathbf{I}}_i$ denote the rendered image for view $i \in \{1, 2\}$, and $\mathbf{M}_i$ be the view-specific mask designed to filter out transient distractors, which will be detailed in Section \ref{sec:masking}. The masked images are given by $\mathbf{I}_i^\mathbf{M} = \mathbf{M}_i \odot \mathbf{I}_i$ and $\tilde{\mathbf{I}}_i^\mathbf{M} = \mathbf{M}_i \odot \tilde{\mathbf{I}}_i$. The reconstruction objective for each view is formulated as:
\begin{equation}
\label{eq:rec_loss}
\mathcal{L}_{i} = (1 - \lambda_{rec}) \|\tilde{\mathbf{I}}_i^\mathbf{M} - \mathbf{I}_i^\mathbf{M}\|_1 + \lambda_{rec} \operatorname{DSSIM}(\tilde{\mathbf{I}}_i^\mathbf{M}, \mathbf{I}_i^\mathbf{M}), \quad i \in \{1, 2\},
\end{equation}
where $\lambda_{rec}$ is a weighting hyperparameter, $\operatorname{DSSIM}$ is the structural dissimilarity index measure \cite{wang2004image}.

\subsection{Conflict-Aware Gradient Harmonization}
\label{sec:gradient_harmonization}

\begin{figure}[t]
\centerline{\includegraphics[width=1.0\textwidth]{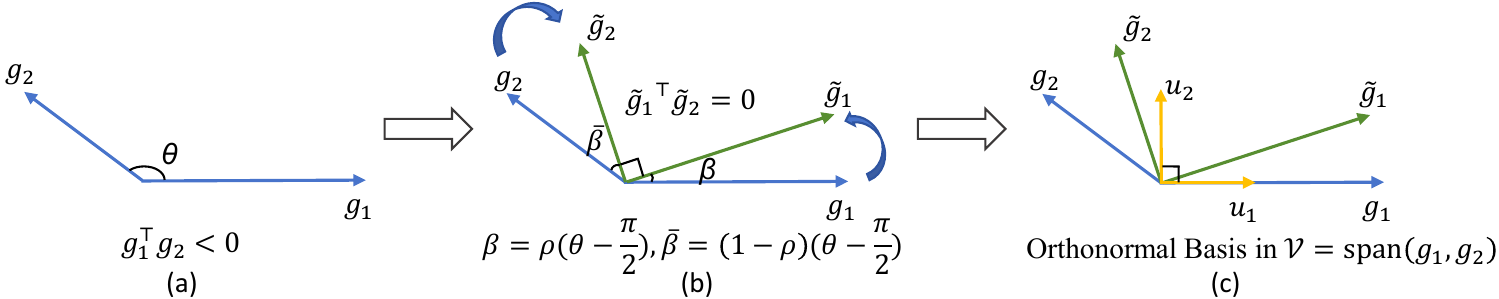}}
\caption{
Illustration of the Cross-View Conflict-Aware Gradient Harmonization process. (a) A gradient conflict is detected when the angle $\theta$ between $g_1$ and $g_2$ is obtuse ($g_1^\top g_2 < 0$). (b) The gradients are mutually rotated within the subspace $\mathcal{V}$ by angles $\beta$ and $\bar{\beta}$ respectively, until they satisfy the orthogonality constraint $\tilde{g}_1^\top \tilde{g}_2 = 0$. (c) Construction of the orthonormal basis $\{u_1, u_2\}$ to facilitate the analytical decomposition and recombination of gradients within the 2D plane.
}
\label{fig:Gradient_Harmonization}
\vspace{-3mm}
\end{figure}

Let $\Theta$ denote the set of optimizable Gaussian parameters. Given the two view-specific losses $\mathcal{L}_1$ and $\mathcal{L}_2$, we compute their corresponding gradients as:
\begin{equation}
g_1 = \nabla_{\Theta} \mathcal{L}_1, \quad
g_2 = \nabla_{\Theta} \mathcal{L}_2.
\end{equation}
A gradient conflict arises when the two gradients form an obtuse angle, i.e., $g_1^\top g_2 < 0$. To mitigate such interference, we harmonize $g_1$ and $g_2$ by mutually rotating them within the subspace $\mathcal{V}=\mathrm{span}(g_1,g_2)$ until the orthogonality constraint $\tilde{g}_1^\top \tilde{g}_2=0$ is satisfied, as shown in Figure \ref{fig:Gradient_Harmonization}. Given the initial angle $\theta = \arccos\left(\frac{g_1^\top g_2}{\|g_1\| \|g_2\|}\right) \in (\frac{\pi}{2}, \pi]$, the total angular correction required to reach orthogonality is $\theta - \frac{\pi}{2}$. We introduce an allocation parameter $\rho \in [0,1]$ to distribute this correction between the two gradients. Specifically, $g_1$ is rotated by $\beta=\rho(\theta-\frac{\pi}{2})$, while $g_2$ is rotated by $\bar{\beta} = (1 - \rho) (\theta - \frac{\pi}{2})$. 
To describe this rotation, we construct an orthonormal basis $\{u_1,u_2\}$ for $\mathcal{V}$, where $u_1$ is aligned with the direction of $g_1$:
\begin{equation}
u_1 = \frac{g_1}{\|g_1\|}, \quad u_2 = \frac{g_2 - (g_2^\top u_1)u_1}{\|g_2 - (g_2^\top u_1)u_1\|}.
\end{equation}
Given that $g_2^\top u_1 = \|g_2\| \cos\theta$ and the magnitude of the orthogonal component is $\|g_2\| \sin\theta$, the expression can be written as:
\begin{equation}
u_2 = \frac{g_2 - \|g_2\| \cos\theta \cdot u_1}{\|g_2\| \sin\theta}.
\end{equation}
Under this basis, the original gradients can be represented as $g_1 = \|g_1\|u_1$ and $g_2 = \|g_2\|(\cos\theta \cdot u_1+ \sin\theta \cdot u_2)$. We then rotate the two gradients while preserving their original magnitudes, yielding the harmonized gradients:
\begin{equation}
\tilde{g}_1 = \|g_1\| (\cos\beta \cdot u_1 + \sin\beta \cdot u_2), \quad \tilde{g}_2 = \|g_2\| (-\sin\beta \cdot u_1 + \cos\beta \cdot u_2).
\end{equation}
Since both harmonized gradients remain in the subspace $\mathcal{V}=\mathrm{span}(g_1,g_2)$, their sum can be expressed as a linear combination of the original gradients:
\begin{equation}
\tilde{g} = \tilde{g}_1 + \tilde{g}_2 = \tau_1 g_1 + \tau_2 g_2 .
\end{equation}
By substituting the expressions of $\tilde{g}_1$ and $\tilde{g}_2$, the coefficients are derived as:
\begin{equation}
\tau_1 = \frac{\sin(\theta - \beta)}{\sin\theta} - \frac{\|g_2\| \cos(\theta - \beta)}{\|g_1\| \sin\theta}, \quad
\tau_2 = \frac{\|g_1\| \sin\beta}{\|g_2\| \sin\theta} + \frac{\cos\beta}{\sin\theta}.
\end{equation}
When no conflict exists, i.e., $g_1^\top g_2 \ge 0$, we set $\tau_1=\tau_2=1$. In practice, we apply the above harmonization independently to each Gaussian attribute. For an attribute $a$, we use its view-specific gradients $g_{1,a}=\nabla_a \mathcal{L}_1$ and $g_{2,a}=\nabla_a \mathcal{L}_2$ to compute the corresponding coefficients $\tau_1^a$ and $\tau_2^a$, where $\tau_i^a$ weights the gradient of attribute $a$ from view $i$.
For geometric attributes $a \in \{\mathbf{x}, \mathbf{r}, \mathbf{s}\}$, since geometry-level conflicts are more likely to induce reconstruction artifacts, we further introduce an attenuation factor when such conflicts are detected:
\begin{equation}
\lambda_{\mathrm{geo}}^{a}
=
\exp\left(-k \cdot \max(0, -\cos\theta_a)\right),
\end{equation}
where $\theta_a$ is the angle between $g_{1,a}$ and $g_{2,a}$, and $k$ controls the attenuation strength. The final harmonized gradient for geometric attributes is then updated by $\tilde{g}^{a} \leftarrow \lambda_{\mathrm{geo}}^{a} \tilde{g}^{a}$.

\begin{figure}[t]
\centerline{\includegraphics[width=0.98\textwidth]{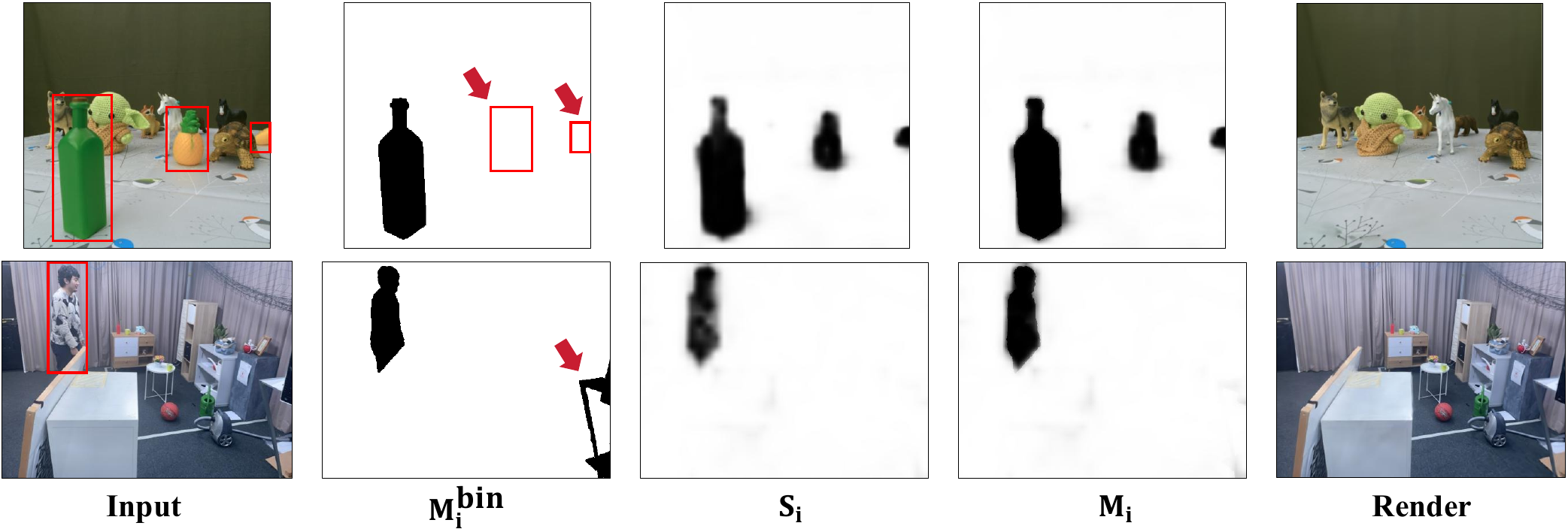}}
\caption{
Visualization of Semantic Consistency-Guided Masking. The SAM-based binary mask $\mathbf{M}_i^{\mathrm{bin}}$ provides a coarse region-level reliability prior, but may miss transient regions or incorrectly suppress reliable areas. 
The learned consistency score $\mathbf{S}_i$ captures pixel-wise inconsistency during training and adaptively refines this binary prior. 
As shown in the highlighted regions, the final mask $\mathbf{M}_i$ suppresses missed transient distractors while correcting erroneous binary-mask decisions, leading to more robust rendering results.
}
\label{fig:mask}
\vspace{-3mm}
\end{figure}

\subsection{Semantic Consistency-Guided Masking}
\label{sec:masking}
Given a ground-truth image $\mathbf{I}_{i}$, we use a frozen DINOv2 \cite{oquab2023dinov2} backbone to extract intermediate features $\mathbf{F}_i \in \mathbb{R}^{C \times H' \times W'}$. A lightweight predictor $f_{\psi}$ maps these semantic priors to a pixel-wise reliability scale $\sigma_i$, where a larger value indicates lower supervision reliability:
\begin{equation}
\sigma_i = \operatorname{softplus}(f_{\psi}(\mathbf{F}_i)+\delta_0),
\end{equation}
where $\delta_0 = \ln(e - 1)$ is an initialization offset ensuring $\sigma_i$ starts near unity. The predicted $\sigma_i$ is defined on the DINOv2 feature grid.
To guide the learning of $\sigma_i$, we construct a semantic consistency-aware residual target from the rendered image $\tilde{\mathbf{I}}_i$ and the ground-truth image $\mathbf{I}_i$:
\begin{equation}
\mathbf{E}_i =
\min\left( 1, \frac{d_{\cos}(\Phi_{\tilde{I}_i}, \Phi_{I_i})}{s_{\mathrm{sem}}} \right)
\odot
\left\|
\mathcal{D}(\tilde{\mathbf{I}}_i)-\mathcal{D}(\mathbf{I}_i)
\right\|_1,
\end{equation}
where $\Phi_{\tilde{I}_i}$ and $\Phi_{I_i}$ are DINOv2 features of the rendered and ground-truth images, respectively, and $\mathcal{D}(\cdot)$ denotes downsampling to the DINOv2 feature resolution. 
The term $d_{\cos}(\cdot, \cdot) = 1 - \frac{\Phi_{\tilde{I}_i} \cdot \Phi_{I_i}}{\|\Phi_{\tilde{I}_i}\| \|\Phi_{I_i}\|}$ computes the cosine distance at each spatial location in the feature space. 
We set $s_{\mathrm{sem}}=0.5$. This formulation increases the response of $\mathbf{E}_i$ only when both the appearance residual and semantic discrepancy are significant, thereby providing a more reliable optimization target.
Based on this, the predictor $f_{\psi}$ is optimized by:
\begin{equation}
\mathcal{L}_{\mathrm{inc}} = \frac{1}{|\Omega|} \sum_{p\in\Omega} \left( \frac{\mathbf{E}_i(p)}{2\sigma_i^2(p)+\epsilon} + \lambda_{\mathrm{inc}}\log(\sigma_i(p)+\epsilon) \right),
\end{equation}
where $\Omega$ denotes the spatial domain of the feature map and $\lambda_{\mathrm{inc}}$ controls the regularization strength. In practice, $\mathbf{E}_i$ is treated as a detached target when optimizing $\mathcal{L}_{\mathrm{inc}}$, so this loss only updates the predictor $f_{\psi}$ and does not propagate gradients to the Gaussian parameters. This formulation encourages the predictor to assign larger $\sigma_i$ to regions characterized by both high photometric residuals and semantic discrepancies. At each training iteration, $\sigma_i$ is first bilinearly upsampled to the image resolution and then converted into a consistency score $\mathbf{S}_i$:
\begin{equation}
\mathbf{S}_i = \exp\left(-\frac{\sigma_i^2}{c_{\sigma}}\right),
\end{equation}
where $c_{\sigma}$ is a scaling hyperparameter. We further construct a SAM-based \cite{kirillov2023segment, li2024segment} binary mask $\mathbf{M}_i^{\mathrm{bin}}$ following AsymGS \cite{li2025robust}, where $\mathbf{M}_i^{\mathrm{bin}}=1$ indicates stable regions and $\mathbf{M}_i^{\mathrm{bin}}=0$ indicates potential transient regions. The learned consistency score $\mathbf{S}_i$ is then combined with this prior to obtain the final reconstruction mask used in Eq.~\ref{eq:rec_loss}:
\begin{equation}
\mathbf{M}_i = \mathbf{M}_i^{\mathrm{bin}} \odot (\mathbf{S}_i)^{\eta_{\mathrm{s}}}
+ (1-\mathbf{M}_i^{\mathrm{bin}}) \odot (\mathbf{S}_i)^{\eta_{\mathrm{t}}},
\end{equation}
where $\eta_{\mathrm{s}}$ and $\eta_{\mathrm{t}}$ control the correction intensity for stable and transient regions, respectively. When $\mathbf{M}_i$ is used in the reconstruction objective, it is detached from the predictor branch, so the reconstruction loss updates the Gaussian parameters without affecting $f_{\psi}$.

\subsection{Conflict-Aware Gaussian Densification and Pruning}
\textbf{Conflict-aware joint densification.} Under our dual-view optimization setting, the single-view densification statistics used in standard 3DGS are insufficient for reliable Gaussian growth. Moreover, directly accumulating raw view-space gradients may make the structural update inconsistent with the harmonized parameter update. We therefore aggregate densification signals from both views, and modulate their view-space gradients using the position-specific conflict-aware coefficients from Section \ref{sec:gradient_harmonization}. Let $\mathcal{V}_i$, $r^{(i)}$, and $\nabla_{\mathbf{p}^{(i)}_{\mathrm{view}}}\mathcal{L}_i$ denote the visible Gaussian set, projected 2D radius, and raw view-space gradient in view $i\in\{1,2\}$, respectively. Given the coefficient $\tau_i^{\mathbf{x}}$ computed for the Gaussian position attribute $\mathbf{x}$, we define the conflict-modulated view-space gradient as:
\begin{equation}
\widetilde{\nabla}_{\mathbf{p}^{(i)}_{\mathrm{view}}}\mathcal{L}_i = \tau_i^{\mathbf{x}}
\nabla_{\mathbf{p}^{(i)}_{\mathrm{view}}}\mathcal{L}_i.
\end{equation}
We then update the joint densification statistics by
\begin{equation}
r_{\max}
\leftarrow
\max\left(
r_{\max},
\mathbb{I}_{\mathcal{V}_1} r^{(1)},
\mathbb{I}_{\mathcal{V}_2} r^{(2)}
\right),
\qquad
\nabla_{\mathrm{accum}}
\leftarrow
\nabla_{\mathrm{accum}}
+
\sum_{i=1}^{2}
\mathbb{I}_{\mathcal{V}_i}
\left\|
\widetilde{\nabla}_{\mathbf{p}^{(i)}_{\mathrm{view}}}\mathcal{L}_i
\right\|.
\end{equation}
The densification criterion is then applied to these joint statistics: the average accumulated gradient determines whether a Gaussian is densified, while $r_{\max}$ determines whether it is cloned or split. 

\textbf{Conflict-guided pruning.} We consider persistent gradient conflicts across views as a potential indicator of unstable Gaussians, which may correspond to floating artifacts. Based on this intuition, we introduce a conflict-guided pruning strategy to suppress such Gaussians during optimization. For the $n$-th Gaussian at iteration $t$, we compute its conflict from the gradients of center position $\mathbf{x}_n$ and opacity $\alpha_n$ in the two training views. Let
$g_{i,n,a}^{(t)}=\nabla_{a_n}\mathcal{L}_i^{(t)}$ denote the gradient of attribute $a\in\{\mathbf{x},\alpha\}$ for Gaussian $n$ from view $i$. The per-Gaussian instantaneous conflict is defined as:
\begin{equation}
C_n^{(t)} = \max_{a\in\{\mathbf{x},\alpha\}} \left[ \max\left( 0,
-\cos\left(g_{1,n,a}^{(t)},g_{2,n,a}^{(t)}\right) \right) \right].
\end{equation}
A larger $C_n^{(t)}$ indicates stronger disagreement between the two views for Gaussian $n$. To reduce the effect of noisy per-iteration gradients, we maintain a per-Gaussian EMA conflict score and periodically decay its opacity:
\begin{equation}
H_n^{(t)} = \gamma H_n^{(t-1)} + (1-\gamma)C_n^{(t)}, \qquad
\alpha_n \leftarrow \alpha_n \cdot \exp\left(-\lambda_{\mathrm{prune}} H_n^{(t)}\right).
\end{equation}
The opacity decay is applied every $100$ iterations. We set $\gamma=0.99$ and $\lambda_{\mathrm{prune}}=0.3$ by default. Gaussians with persistent conflicts gradually lose opacity and are eventually removed by the standard Gaussian pruning step, thereby suppressing floating artifacts while preserving optimization stability.

\begin{table}[t]
\centering
\caption{Quantitative results on the NeRF On-the-go dataset. The \textbf{best} and \underline{second-best} values in each column are highlighted in bold and underlined, respectively.}
\label{tab:onthego-comparison}
\renewcommand{\arraystretch}{0.95}
\setlength{\tabcolsep}{6pt}
\resizebox{\textwidth}{!}{
\begin{tabular}{l|ccc|ccc|ccc}
\toprule
Scene & \multicolumn{3}{c|}{High Occlusion} & \multicolumn{3}{c|}{Medium Occlusion} & \multicolumn{3}{c}{Low Occlusion} \\
\midrule
Method 
 & PSNR$\uparrow$ & SSIM$\uparrow$ & LPIPS$\downarrow$
 & PSNR$\uparrow$ & SSIM$\uparrow$ & LPIPS$\downarrow$
 & PSNR$\uparrow$ & SSIM$\uparrow$ & LPIPS$\downarrow$ \\
\midrule
RobustNeRF~\cite{sabour2023robustnerf}
 & 20.60 & 0.602 & 0.379
 & 21.72 & 0.741 & 0.248
 & 16.60 & 0.407 & 0.480 \\
NeRF On-the-go~\cite{ren2024nerf}
 & 22.37 & 0.753 & 0.212
 & 22.50 & 0.780 & 0.205
 & 20.13 & 0.627 & 0.287 \\
3DGS~\cite{kerbl20233d}
 & 19.03 & 0.649 & 0.340
 & 19.19 & 0.709 & 0.220
 & 19.68 & 0.649 & 0.199 \\
Mip-Splatting~\cite{yu2024mip}
 & 19.25 & 0.664 & 0.333
 & 19.73 & 0.684 & 0.279
 & 20.03 & 0.661 & 0.195 \\
GS-W~\cite{zhang2024gaussian}
 & 18.52 & 0.645 & 0.335
 & 21.04 & 0.737 & 0.208
 & 19.75 & 0.660 & 0.287 \\
WildGaussian~\cite{kulhanek2024wildgaussians}
 & 23.03 & 0.771 & 0.172
 & 22.80 & 0.811 & 0.092
 & 20.62 & 0.658 & 0.235 \\
SpotLessSplats~\cite{sabour2025spotlesssplats}
 & 21.92 & 0.710 & 0.222
 & 22.79 & 0.817 & 0.162
 & 20.02 & 0.596 & 0.276 \\
HybridGS~\cite{lin2025hybridgs}
 & 23.05 & 0.768 & 0.204
 & 23.51 & 0.830 & 0.160
 & 21.42 & 0.684 & 0.268 \\
DroneSplat~\cite{tang2025dronesplat}
 & 23.52 & 0.810 & \underline{0.136}
 & 23.27 & 0.813 & 0.102
 & 21.53 & 0.697 & \textbf{0.162} \\
AsymGS (GS-GS)~\cite{li2025robust}
 & \underline{24.34} & \underline{0.825} & 0.150
 & \underline{24.56} & \underline{0.872} & \underline{0.090}
 & \underline{21.91} & \underline{0.728} & 0.189 \\
AsymGS (EMA-GS)~\cite{li2025robust}
 & 24.12 & 0.818 & 0.154
 & 24.32 & 0.864 & \underline{0.090}
 & 21.77 & 0.722 & \textbf{0.162} \\
\midrule
\textbf{Ours}
 & \textbf{24.67} & \textbf{0.844} & \textbf{0.123}
 & \textbf{25.09} & \textbf{0.879} & \textbf{0.084}
 & \textbf{22.17} & \textbf{0.730} & \underline{0.164} \\
\bottomrule
\end{tabular}
}
\vspace{-3mm}
\end{table}

\section{Experiment}

\subsection{Experimental Setup}
\textbf{Datasets.} We evaluate our method on three widely used in-the-wild benchmarks: PhotoTourism~\cite{martin2021nerf}, NeRF On-the-go~\cite{ren2024nerf}, and RobustNeRF~\cite{sabour2023robustnerf}. PhotoTourism consists of internet landmark photo collections, featuring substantial appearance variations across time, weather, and seasons, as well as transient occluders such as pedestrians, vehicles, and scaffolding. Following prior works~\cite{chen2022hallucinated, martin2021nerf, kulhanek2024wildgaussians}, we evaluate on three scenes: Brandenburg Gate, Sacre Coeur, and Trevi Fountain. NeRF On-the-go contains handheld video sequences with dynamic occlusions, and we evaluate on all six public scenes spanning low-, medium-, and high-occlusion settings. RobustNeRF assesses reconstruction robustness under severe transient distractions, including challenging occlusions and inconsistent observations across views. We evaluate on all four scenes: Statue, Android, Crab, and Yoda.

\textbf{Evaluation Metrics.} Following previous methods \cite{sabour2023robustnerf,ren2024nerf,kerbl20233d,yu2024mip}, we mainly evaluate the rendering quality with three standard metrics: Peak Signal-to-Noise Ratio (PSNR), Structural Similarity Index Measure (SSIM) \cite{wang2004image}, and Learned Perceptual Image Patch Similarity (LPIPS) \cite{zhang2018unreasonable}. 

\textbf{Implementation Details.}
Our Gaussian model is built upon Mip-Splatting~\cite{yu2024mip}. We follow AsymGS~\cite{li2025robust} for the overall densification and pruning schedule, while adapting the optimization settings to our framework.
Relative to the default settings, the learning rates of Gaussian attributes are uniformly scaled by $0.7$ on NeRF On-the-go and RobustNeRF, and by $0.5$ on PhotoTourism. The reconstruction loss weight $\lambda_{\mathrm{rec}}$ is set to $0.25$. For Semantic Consistency-Guided Masking, we adopt a $5{,}000$-iteration warm-up, during which only the prior binary mask is used for reconstruction. The learning rate of the $\sigma_i$ predictor is set to $0.001$, the regularization weight $\lambda_{\mathrm{inc}}$ is set to $0.5$, and the scaling hyperparameter $c_{\sigma}$ for the consistency score $\mathbf{S}_i$ is set to $0.2$. The mask correction parameters are set to $\eta_{\mathrm{s}}=1.2$ and $\eta_{\mathrm{t}}=3$ for stable and transient regions, respectively. For Conflict-Aware Gradient Harmonization, the geometric attenuation strength is set to $k=0.5$.

\begin{table}[t]
\caption{Quantitative comparison on the RobustNeRF dataset. The \textbf{best} and \underline{second-best} results in each column are highlighted in bold and underlined, respectively.}
\label{tab:robustnerf-comparison}
\centering
\renewcommand{\arraystretch}{0.95}
\setlength{\tabcolsep}{4pt}
\resizebox{\textwidth}{!}{%
\begin{tabular}{l|ccc|ccc|ccc|ccc}
\toprule
Scene  & \multicolumn{3}{c|}{Statue}  & \multicolumn{3}{c|}{Android} & \multicolumn{3}{c|}{Yoda} & \multicolumn{3}{c}{Crab} \\
\midrule
Method
 & PSNR$\uparrow$ & SSIM$\uparrow$ & LPIPS$\downarrow$
 & PSNR$\uparrow$ & SSIM$\uparrow$ & LPIPS$\downarrow$
 & PSNR$\uparrow$ & SSIM$\uparrow$ & LPIPS$\downarrow$
 & PSNR$\uparrow$ & SSIM$\uparrow$ & LPIPS$\downarrow$ \\
\midrule
RobustNeRF~\cite{sabour2023robustnerf}
 & 20.60 & 0.760 & 0.150
 & 23.28 & 0.750 & 0.130
 & 29.78 & 0.820 & 0.150
 & -     & -     & -      \\
NeRF On-the-go~\cite{ren2024nerf}
 & 21.58 & 0.770 & 0.240
 & 23.50 & 0.750 & 0.210
 & 29.96 & 0.830 & 0.240
 & -     & -     & -      \\
3DGS~\cite{kerbl20233d}
 & 21.02 & 0.810 & 0.160
 & 23.11 & 0.810 & 0.130
 & 26.33 & 0.910 & 0.140
 & 29.74 & -     & -      \\
Mip-Splatting~\cite{yu2024mip}
 & 22.08 & 0.860 & 0.135
 & 23.45 & 0.801 & 0.106
 & 27.96 & 0.933 & 0.136
 & 29.18 & 0.929 & 0.129  \\
GS-W~\cite{zhang2024gaussian}
 & 21.99 & 0.862 & 0.102
 & 24.23 & 0.824 & 0.090
 & 32.74 & 0.957 & 0.084
 & 33.22 & 0.952 & 0.088  \\
WildGaussian~\cite{kulhanek2024wildgaussians}
 & 23.25 & 0.886 & 0.105
 & 24.57 & 0.827 & 0.085
 & 32.84 & 0.956 & 0.091
 & 32.81 & 0.952 & 0.092  \\
SpotLessSplats~\cite{sabour2025spotlesssplats}
 & 22.54 & 0.840 & 0.130
 & 25.05 & 0.850 & 0.090
 & 33.66 & 0.960 & 0.100
 & 34.43 & -     & -      \\
HybridGS~\cite{lin2025hybridgs}
 & 22.93 & 0.870 & 0.100
 & 25.15 & 0.850 & 0.070
 & 35.32 & 0.960 & \textbf{0.070}
 & 35.17 & 0.960 & 0.080  \\
AsymGS (GS-GS)~\cite{li2025robust}
& 23.47 & \textbf{0.894} & 0.097
& \underline{25.61} & \textbf{0.857} & 0.071
& \underline{37.18} & \textbf{0.969} & \underline{0.074}
& \underline{36.18} & \textbf{0.964} & \underline{0.078}  \\
AsymGS (EMA-GS)~\cite{li2025robust}
 & \underline{23.49} & 0.890 & \underline{0.096}
 & 25.47 & 0.849 & \underline{0.068}
 & 36.50 & \underline{0.967} & 0.077
 & 35.60 & \underline{0.961} & 0.080  \\
\midrule
\textbf{Ours}
 & \textbf{23.95} & \underline{0.893} & \textbf{0.095}
 & \textbf{26.09}    & \textbf{0.857} & \textbf{0.067}
 & \textbf{37.29}    & \textbf{0.969} & \underline{0.074}
 & \textbf{36.50}    & \textbf{0.964} & \textbf{0.077}  \\
\bottomrule
\end{tabular}%
}
\vspace{-2mm}
\end{table}

\begin{figure}[t]
\centerline{\includegraphics[width=1.0\textwidth]{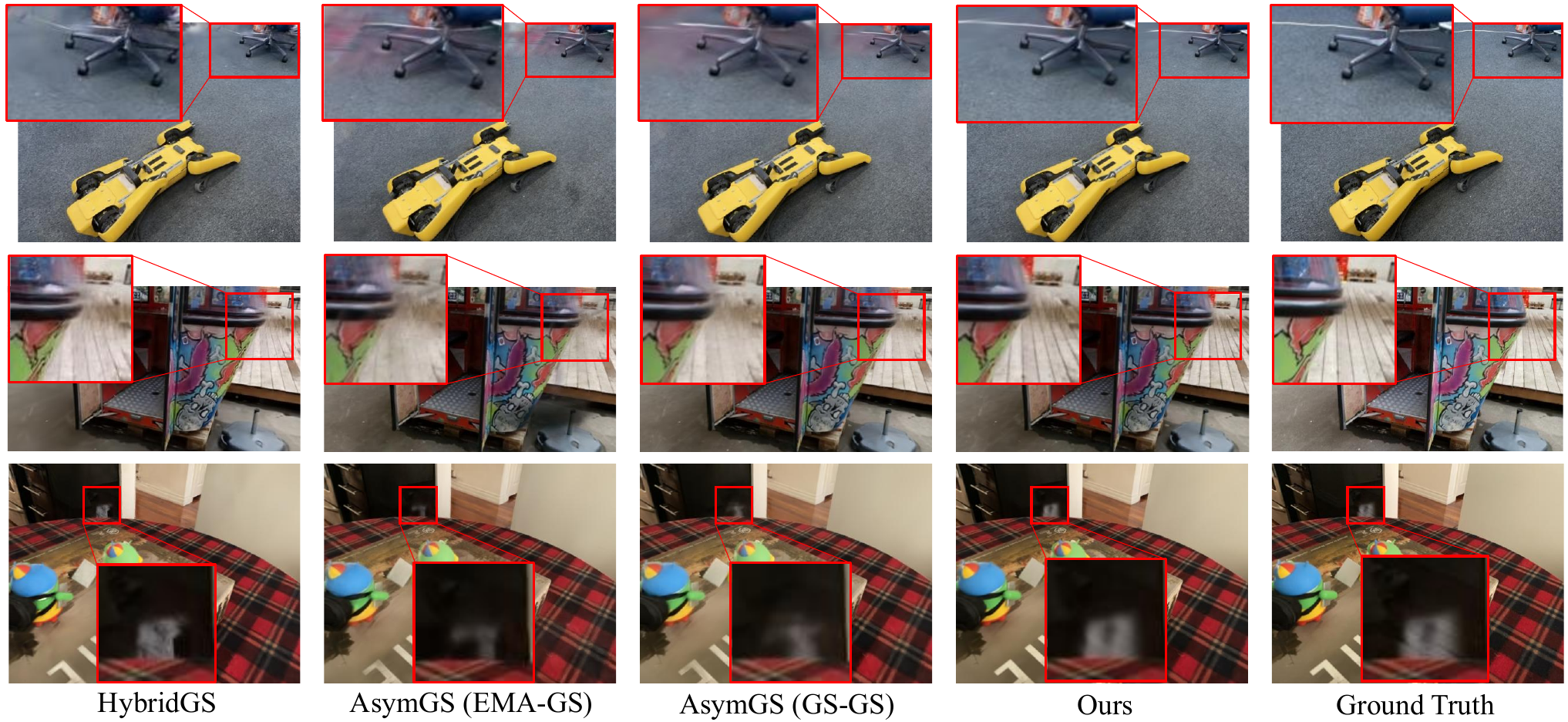}}
\caption{
Qualitative comparisons on challenging in-the-wild scenes. Our method suppresses transient artifacts more effectively than current state-of-the-art methods, yielding cleaner renderings and more faithful scene details.
}
\label{fig:qualitative_result}
\vspace{-2mm}
\end{figure}

\subsection{Comparison with State-of-the-Art Methods}
As shown in Tables~\ref{tab:onthego-comparison}, \ref{tab:robustnerf-comparison}, and \ref{tab:phototourism-comparison}, our method consistently achieves superior reconstruction quality over both NeRF-based and Gaussian-based baselines across all evaluated in-the-wild datasets, including NeRF On-the-go, RobustNeRF, and PhotoTourism. These results demonstrate the robustness and generality of our approach under challenging transient distractors and cross-view inconsistencies.
Figure~\ref{fig:qualitative_result} further provides visual comparisons with representative state-of-the-art methods. The qualitative results show that our method effectively suppresses floating artifacts, leading to cleaner renderings and more faithful reconstructions.

\begin{table}[t]
\centering
\caption{Quantitative comparison on the PhotoTourism dataset. The \textbf{best} and \underline{second-best} results in each column are highlighted in bold and underlined, respectively.}
\label{tab:phototourism-comparison}
\renewcommand{\arraystretch}{0.95}
\setlength{\tabcolsep}{4pt}
\resizebox{\textwidth}{!}{%
\begin{tabular}{l|ccc|ccc|ccc}
\toprule
Scene
 & \multicolumn{3}{c|}{Brandenburg Gate}
 & \multicolumn{3}{c|}{Sacre Coeur}
 & \multicolumn{3}{c}{Trevi Fountain} \\
\midrule
Method
 & PSNR$\uparrow$ & SSIM$\uparrow$ & LPIPS$\downarrow$
 & PSNR$\uparrow$ & SSIM$\uparrow$ & LPIPS$\downarrow$
 & PSNR$\uparrow$ & SSIM$\uparrow$ & LPIPS$\downarrow$ \\
\midrule
NeRF~\cite{mildenhall2021nerf}
 & 18.90 & 0.882 & 0.138
 & 15.60 & 0.846 & 0.163
 & 16.14 & 0.696 & 0.282 \\
3DGS~\cite{kerbl20233d}
 & 19.37 & 0.880 & 0.141
 & 17.44 & 0.835 & 0.204
 & 17.58 & 0.709 & 0.266 \\
Mip-Splatting~\cite{yu2024mip}
 & 20.01 & 0.877 & 0.166
 & 17.54 & 0.831 & 0.203
 & 17.36 & 0.684 & 0.319 \\
NeRF-W~\cite{martin2021nerf}
 & 24.17 & 0.891 & 0.152
 & 19.20 & 0.803 & 0.192
 & 18.97 & 0.698 & 0.265 \\
Ha-NeRF~\cite{chen2022hallucinated}
 & 24.04 & 0.887 & 0.139
 & 20.02 & 0.801 & 0.171
 & 20.18 & 0.691 & 0.223 \\
K-Planes~\cite{fridovich2023k}
 & 25.49 & 0.879 & 0.224
 & 20.61 & 0.774 & 0.265
 & 22.67 & 0.714 & 0.317 \\
RefinedFields~\cite{kassab2023refinedfields}
 & 26.64 & 0.886 & -
 & 22.26 & 0.817 & -
 & 23.42 & 0.737 & - \\
GS-W~\cite{zhang2024gaussian}
 & 23.51 & 0.897 & 0.166
 & 19.39 & 0.825 & 0.211
 & 20.06 & 0.723 & 0.274 \\
WildGaussian~\cite{kulhanek2024wildgaussians}
 & 27.77 & 0.927 & 0.133
 & 22.56 & 0.859 & 0.177
 & 23.63 & 0.766 & 0.228 \\
AsymGS (GS-GS)~\cite{li2025robust}
 & \underline{28.56} & \underline{0.938} & \underline{0.109}
 & \underline{23.78} & \underline{0.887} & \textbf{0.139}
 & \underline{24.52} & \underline{0.790} & \underline{0.202} \\
AsymGS (EMA-GS)~\cite{li2025robust}
 & 28.50 & 0.937 & 0.115
 & 23.37 & 0.882 & 0.150
 & 23.85 & 0.775 & 0.242 \\
\midrule
Ours
 & \textbf{29.08} & \textbf{0.945} & \textbf{0.102}
 & \textbf{24.58} & \textbf{0.892} & \underline{0.143}
 & \textbf{24.97} & \textbf{0.807} & \textbf{0.201} \\
\bottomrule
\end{tabular}%
}
\vspace{-3mm}
\end{table}

\begin{table}[t]
\centering
\caption{
Ablation study on the NeRF On-the-go dataset. We compare the full model with recent state-of-the-art methods, examine the contribution of each component in the proposed dual-view framework, and further report single-view variants to isolate the effect of dual-view optimization.
}
\label{tab:onthego-ablation}
\renewcommand{\arraystretch}{1.15}
\setlength{\tabcolsep}{6pt}
\resizebox{1\textwidth}{!}{
\begin{tabular}{l|cccccc}
\toprule
Method 
 & PSNR$\uparrow$ & SSIM$\uparrow$ & LPIPS$\downarrow$
 & Training Time (h)  & GPU Memory (GB)  &  Points (K)\\
\midrule
\multicolumn{7}{c}{\textbf{Baselines}} \\
\midrule
AsymGS (GS-GS)~\cite{li2025robust}
 & 23.61 & 0.809 & 0.143
 & 0.41  & 3.73 & 546 \\
AsymGS (EMA-GS)~\cite{li2025robust}
 & 23.40 & 0.801 & 0.135
 & 0.27  & 2.82 & 531 \\
\midrule
\multicolumn{7}{c}{\textbf{Dual-view Setting}} \\
\midrule
\textbf{Ours}
 & 23.98 & 0.818 & 0.123
 & 0.33 & 3.97 & 344 \\
\textbf{Ours w/o $\mathbf{M}_i$}
 & 22.43 & 0.770 & 0.169
 & 0.15 & 3.21 & 444 \\
\textbf{Ours w/o $\mathbf{S}_i$ (only $\mathbf{M}_i^{\mathrm{bin}}$)}
 & 23.46 & 0.806 & 0.134
 & 0.16  & 3.38 & 430 \\
\textbf{Ours w/o $\mathbf{M}_i^{\mathrm{bin}}$ (only $\mathbf{S}_i$)}
 & 23.52 & 0.799 & 0.139
 & 0.30  & 3.92 & 396 \\
\textbf{Ours w/o Gradient Harmonization}
 & 23.50 & 0.802 & 0.131
 & 0.33  & 3.96 & 471 \\
\textbf{Ours w/o Conf.-Aware Dens. \& Prun.}
 & 23.84 & 0.814 & 0.128
 & 0.30  & 3.98 & 413 \\
\midrule
\multicolumn{7}{c}{\textbf{Single-view Setting}} \\
\midrule
\textbf{Ours (single-view)}
 & 23.49 & 0.801 & 0.139
 & 0.21 & 3.29 & 393 \\
\textbf{Ours (single-view) w/o $\mathbf{M}_i$}
 & 22.27 & 0.776 & 0.166
 & 0.14  & 3.08  & 467 \\
\textbf{Ours (single-view) w/o $\mathbf{S}_i$ (only $\mathbf{M}_i^{\mathrm{bin}}$)}
 & 23.25 & 0.803 & 0.133
 & 0.15  & 3.08 & 462 \\
\textbf{Ours (single-view) w/o $\mathbf{M}_i^{\mathrm{bin}}$ (only $\mathbf{S}_i$)}
 & 23.43 & 0.804 & 0.136
 & 0.20  & 3.26 & 399 \\
\bottomrule
\end{tabular}%
}
\vspace{-3mm}
\end{table}

\subsection{Ablation Study}
\textbf{Gradient harmonization and Gaussian structure refinement.}
As shown in Table~\ref{tab:onthego-ablation}, our full dual-view framework achieves the best reconstruction quality among all variants. Compared with their single-view counterparts, the dual-view variants generally improve reconstruction performance under different masking configurations, especially in terms of PSNR, including w/o $\mathbf{M}_i$, w/o $\mathbf{S}_i$ (only $\mathbf{M}_i^{\mathrm{bin}}$), and w/o $\mathbf{M}_i^{\mathrm{bin}}$ (only $\mathbf{S}_i$). Specifically, dual-view optimization improves PSNR from 22.27 to 22.43 without $\mathbf{M}_i$, from 23.25 to 23.46 when only $\mathbf{M}_i^{\mathrm{bin}}$ is used, and from 23.43 to 23.52 when only $\mathbf{S}_i$ is used. These results indicate that Gradient Harmonization helps exploit two-view optimization signals while mitigating their potential conflicts.

We further ablate the two conflict-aware components. Removing conflict-aware Gradient Harmonization reduces PSNR from 23.98 to 23.50 and SSIM from 0.818 to 0.802, indicating that directly combining two-view gradients without conflict reconciliation leads to less stable optimization. Removing conflict-aware densification and pruning also degrades performance, reducing PSNR to 23.84. Although this variant remains competitive, it uses more Gaussians than the full model (413K vs. 344K). This suggests that conflict-aware structure refinement, i.e., densification and pruning, not only improves reconstruction quality but also produces a more compact Gaussian representation by suppressing unstable primitives and avoiding redundant growth.

\textbf{Design for masking.} Removing the final mask $\mathbf{M}_i$ clearly degrades performance, dropping PSNR from 23.98 to 22.43 in the dual-view setting and from 23.49 to 22.27 in the single-view setting. These results show that using masks to suppress unreliable transient regions is crucial for robust in-the-wild reconstruction. We further evaluate the two components of $\mathbf{M}_i$ separately. In the dual-view setting, using only the SAM-based binary mask $\mathbf{M}_i^{\mathrm{bin}}$ achieves 23.46 PSNR and 0.134 LPIPS, while using only the learned consistency score $\mathbf{S}_i$ achieves 23.52 PSNR and 0.139 LPIPS. Similar results are observed in the single-view setting, where the two variants obtain 23.25 and 23.43 PSNR, respectively, both outperforming the no-mask variant. These results show that both components are effective, while their combination consistently achieves the best performance, indicating that the binary mask and learned consistency score provide complementary cues for suppressing unreliable regions.

\section{Conclusion}
In this work, we present a novel conflict-aware 3D Gaussian Splatting framework for robust in-the-wild scene reconstruction. We first introduce Semantic Consistency-Guided Masking to adaptively refine prior masks and effectively suppress unreliable supervision. Since masking alone cannot fully resolve cross-view optimization conflicts, we further propose dual-view Conflict-Aware Gradient Harmonization to reconcile view-specific gradients into an orthogonal configuration. Conflict-aware densification and pruning are also developed to stabilize Gaussian structure optimization by guiding reliable growth and removing persistently conflicting primitives. Extensive experiments on three challenging in-the-wild benchmarks demonstrate the effectiveness and robustness of our approach.


\bibliographystyle{plain}
\bibliography{main}






\newpage
\appendix

\section{Additional Results}
\begin{figure}[t]
\centerline{\includegraphics[width=0.8\textwidth]{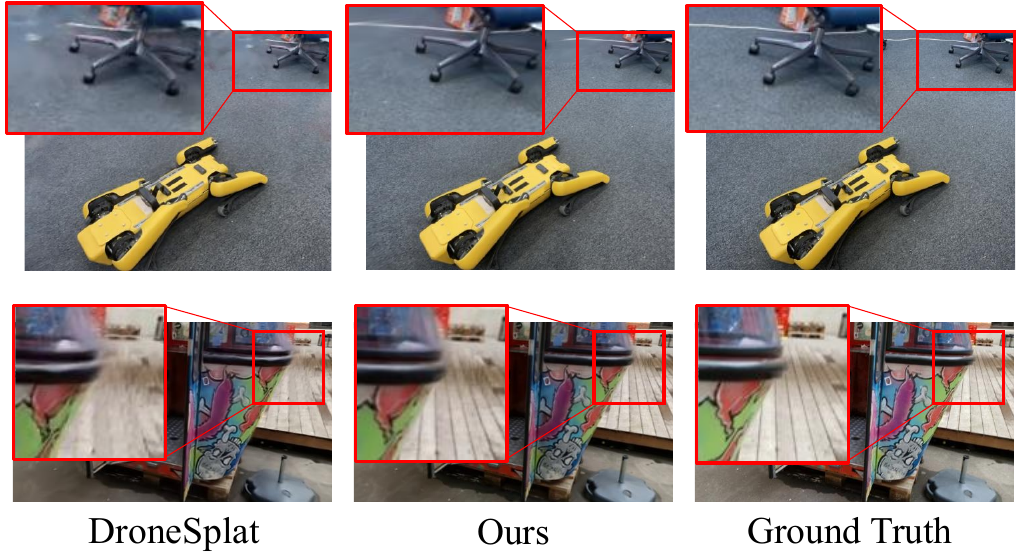}}
\caption{
Additional qualitative comparisons with DroneSplat~\cite{tang2025dronesplat}.
}
\label{fig:additional_qualitative_result1}
\vspace{0mm}
\end{figure}

\begin{figure}[t]
\centerline{\includegraphics[width=1.0\textwidth]{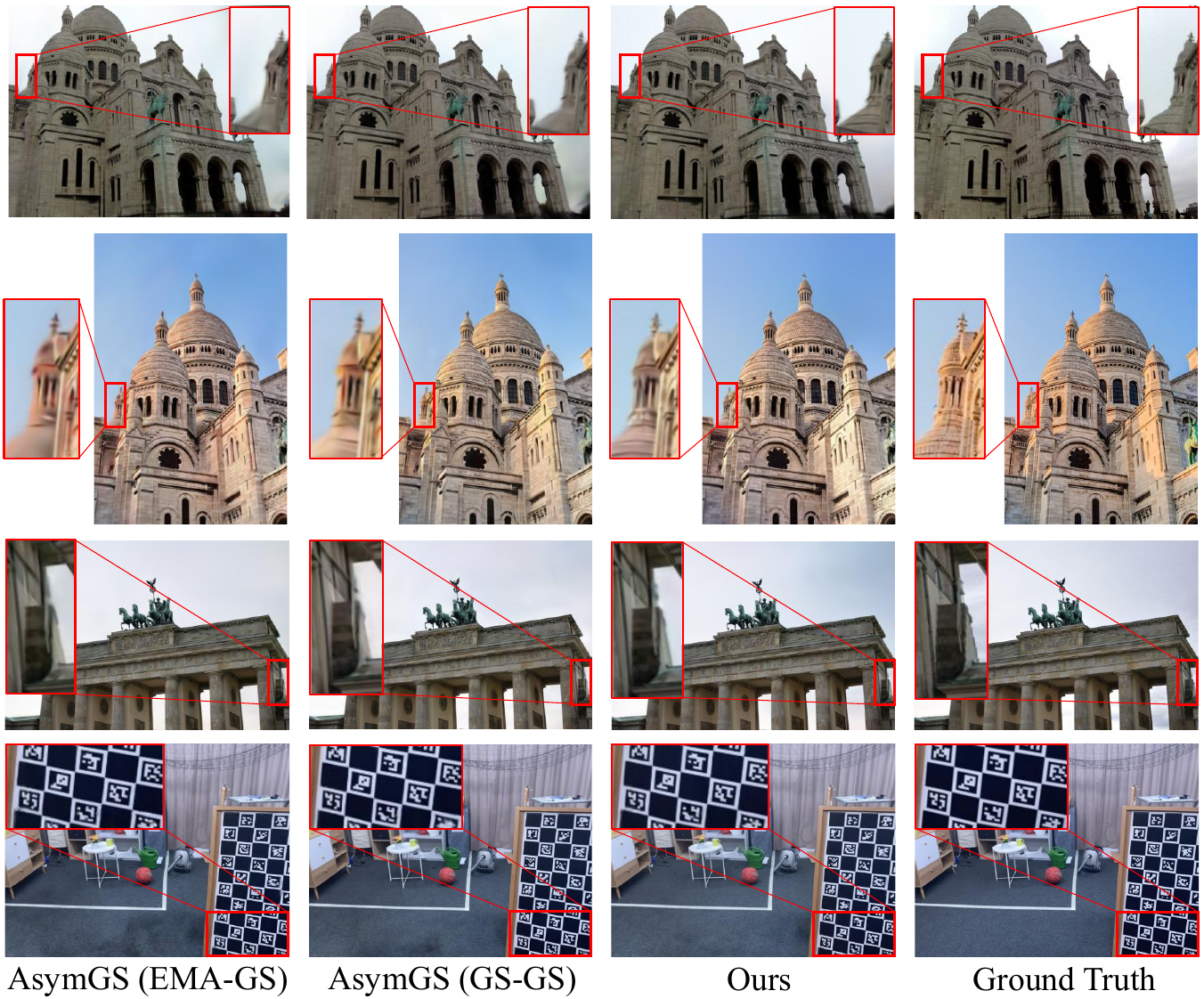}}
\caption{
Additional qualitative comparisons with AsymGS~\cite{li2025robust}.
}
\label{fig:additional_qualitative_result2}
\vspace{0mm}
\end{figure}

Complete per-scene results on the NeRF On-the-go dataset are reported in Table~\ref{tab:onthego-per-scene}, and additional qualitative comparisons are provided in Figures \ref{fig:additional_qualitative_result1} and \ref{fig:additional_qualitative_result2}.

\begin{table}[t]
\centering
\caption{Per-scene quantitative results on On-the-go dataset. The \textbf{best} and \underline{second-best} values in each column are highlighted in bold and underlined, respectively.}
\label{tab:onthego-per-scene}
\renewcommand{\arraystretch}{1.15}
\setlength{\tabcolsep}{6pt}
\resizebox{\textwidth}{!}{
\begin{tabular}{l|ccc|ccc|ccc|ccc|ccc|ccc}
\toprule
Scene & \multicolumn{3}{c|}{Mountain} & \multicolumn{3}{c|}{Fountain} & \multicolumn{3}{c|}{Corner}& \multicolumn{3}{c|}{Patio} & \multicolumn{3}{c|}{Spot} & \multicolumn{3}{c}{Patio-High} \\
\midrule
Method 
 & PSNR$\uparrow$ & SSIM$\uparrow$ & LPIPS$\downarrow$
 & PSNR$\uparrow$ & SSIM$\uparrow$ & LPIPS$\downarrow$
 & PSNR$\uparrow$ & SSIM$\uparrow$ & LPIPS$\downarrow$
 & PSNR$\uparrow$ & SSIM$\uparrow$ & LPIPS$\downarrow$
 & PSNR$\uparrow$ & SSIM$\uparrow$ & LPIPS$\downarrow$
 & PSNR$\uparrow$ & SSIM$\uparrow$ & LPIPS$\downarrow$ \\
\midrule
RobustNeRF~\cite{sabour2023robustnerf}
 & 17.54 & 0.496 & 0.383
 & 15.65 & 0.318 & 0.576
 & 23.04 & 0.764 & 0.244
 & 20.39 & 0.718 & 0.251
 & 20.65 & 0.625 & 0.391
 & 20.54 & 0.578 & 0.366 \\
NeRF On-the-go~\cite{ren2024nerf}
 & 20.15 & 0.644 & 0.259
 & 20.11 & 0.609 & 0.314
 & 24.22 & 0.806 & 0.190
 & 20.78 & 0.754 & 0.219
 & 23.33 & 0.787 & 0.189
 & 21.41 & 0.718 & 0.235 \\
3DGS~\cite{kerbl20233d}
 & 19.40 & 0.638 & 0.213
 & 19.96 & 0.659 & 0.185
 & 20.90 & 0.713 & 0.241
 & 17.48 & 0.704 & 0.199
 & 20.77 & 0.693 & 0.316
 & 17.29 & 0.604 & 0.363 \\
Mip-Splatting~\cite{yu2024mip}
 & 19.86 & 0.649 & 0.200
 & 20.19 & 0.672 & 0.189
 & 21.15 & 0.728 & 0.230 
 & 18.31 & 0.639 & 0.328
 & 20.18 & 0.689 & 0.338
 & 18.31 & 0.639 & 0.328 \\
GS-W~\cite{zhang2024gaussian}
 & 19.92 & 0.560 & 0.291
 & 20.19 & 0.589 & 0.279
 & 23.72 & 0.785 & 0.132
 & 19.10 & 0.691 & 0.163
 & 22.42 & 0.635 & 0.309
 & 21.21 & 0.649 & 0.260 \\
WildGaussian~\cite{kulhanek2024wildgaussians}
 & 20.43 & 0.653 & 0.255
 & 20.81 & 0.662 & 0.215
 & 24.16 & 0.822 & \textbf{0.045}
 & 21.44 & 0.800 & 0.138
 & 23.82 & 0.816 & 0.138
 & 22.23 & 0.725 & 0.206 \\
SpotLessSplats~\cite{sabour2025spotlesssplats}
 & 19.84 & 0.580 & 0.294
 & 20.19 & 0.612 & 0.258
 & 24.03 & 0.795 & 0.258
 & 21.55 & 0.838 & \textbf{0.065}
 & 23.52 & 0.756 & 0.185
 & 20.31 & 0.664 & 0.259 \\
HybridGS~\cite{lin2025hybridgs}
 & 21.73 & 0.693 & 0.284
 & 21.11 & 0.674 & 0.252
 & 25.03 & 0.847 & 0.151
 & 21.98 & 0.812 & 0.169
 & 24.33 & 0.794 & 0.196
 & 21.77 & 0.741 & 0.211 \\
DroneSplat~\cite{tang2025dronesplat}
 & 21.45 & 0.694 & \textbf{0.158}
 & 21.60 & 0.699 & 0.165
 & 24.65 & 0.814 & 0.098
 & 21.88 & 0.812 & 0.106
 & 24.44 & 0.827 & \textbf{0.094}
 & 22.60 & 0.792 & 0.177 \\
AsymGS (GS-GS)~\cite{li2025robust}
 & \underline{22.00} & \underline{0.740} & 0.199
 & \underline{21.83} & \underline{0.717} & 0.180
 & \underline{26.15} & \underline{0.885} & 0.085
 & 22.97 & \underline{0.860} & 0.096
 & \underline{25.52} & \underline{0.854} & 0.135
 & \underline{23.17} & 0.796 & 0.164 \\
AsymGS (EMA-GS)~\cite{li2025robust}
 & 21.93 & 0.735 & \underline{0.162}
 & 21.61 & 0.709 & \underline{0.162}
 & 25.77 & 0.876 & 0.089
 & 22.87 & 0.853 & 0.091
 & 25.09 & 0.839 & 0.152
 & 23.14 & \underline{0.797} & \underline{0.156} \\
\midrule
\textbf{Ours}
 & \textbf{22.21} & \textbf{0.741} & 0.177
 & \textbf{22.12} & \textbf{0.718} & \textbf{0.151}
 & \textbf{26.58} & \textbf{0.886} & \underline{0.081} 
& \textbf{23.59} & \textbf{0.871} & \underline{0.087}
 & \textbf{25.91} & \textbf{0.879} & \underline{0.096}
 & \textbf{23.43} & \textbf{0.809} & \textbf{0.149} \\
\bottomrule
\end{tabular}
}
\end{table}

\section{Limitations}
We observe that Conflict-Aware Gradient Harmonization can be sensitive to mask quality. As shown in Table~\ref{tab:onthego-ablation}, when the mask is less reliable, e.g., w/o $\mathbf{M}_i$ and w/o $\mathbf{M}_i^{\mathrm{bin}}$ (only $\mathbf{S}_i$), the dual-view setting slightly underperforms its single-view counterpart on SSIM or LPIPS in a few cases. We hypothesize that inaccurate masks may introduce unreliable regions into the two-view optimization process. In this case, harmonizing gradients from two views can lead to a compromise between inconsistent signals, which may slightly weaken fine structural details. Nevertheless, the dual-view setting still brings clear PSNR improvements, indicating that it improves the overall reconstruction fidelity even when local perceptual details are mildly affected. This also highlights the importance of accurate inconsistency masking for fully exploiting cross-view gradient harmonization.

\section{Additional Analysis on Cross-View Gradient Harmonization}
\label{app:conflict_analysis}
This section provides additional analysis and derivation details for the proposed two-view gradient harmonization strategy. 
We first discuss the delayed nature of conflicts in single-view optimization in Section~\ref{app:single_view_conflicts}. 
We then provide a detailed derivation of the harmonized gradients in Section~\ref{app:harmonized_gradient_derivation}, followed by an orthogonality verification in Section~\ref{app:orthogonality_verification}. 
Next, we derive how the harmonized update can be reformulated in the original gradient basis in Section~\ref{app:gradient_basis_recombination}. 
Finally, we analyze the expected pairwise conflict coverage of random two-view sampling in Section~\ref{app:expected_conflict_coverage}.

\subsection{Implicit and Delayed Conflicts in Single-View Optimization}
\label{app:single_view_conflicts}
Assume the training set contains $N$ views. 
For the $i$-th view, let its reconstruction objective be $\mathcal{L}_i(\Theta)$, where $\Theta$ denotes the shared optimizable Gaussian parameters. 
The corresponding view-specific gradient is
\begin{equation}
    g_i = \nabla_{\Theta} \mathcal{L}_i(\Theta).
\end{equation}
For two different views $i$ and $j$, a pairwise gradient conflict occurs when
\begin{equation}
    g_i^\top g_j < 0.
\end{equation}
This condition indicates that the two views prefer incompatible update directions for the shared Gaussian parameters.

Conventional 3DGS optimization samples one view at each iteration. 
If view $i$ is sampled, the parameter update is
\begin{equation}
    \Theta^{+} = \Theta - \eta g_i,
\end{equation}
where $\eta$ is the learning rate. 
Since only $g_i$ is computed, the current iteration cannot observe whether this update is compatible with other views. 
However, the update may still increase the loss of another view $j$. 
Using a first-order Taylor expansion around $\Theta$, we have
\begin{equation}
    \mathcal{L}_j(\Theta^{+})
    =
    \mathcal{L}_j(\Theta - \eta g_i)
    \approx
    \mathcal{L}_j(\Theta)
    -
    \eta \nabla_{\Theta}\mathcal{L}_j(\Theta)^\top g_i.
\end{equation}
Since $\nabla_{\Theta}\mathcal{L}_j(\Theta)=g_j$, this becomes
\begin{equation}
    \mathcal{L}_j(\Theta^{+})
    \approx
    \mathcal{L}_j(\Theta)
    -
    \eta g_j^\top g_i.
\end{equation}
If $g_i^\top g_j < 0$, then
\begin{equation}
    -\eta g_j^\top g_i > 0,
\end{equation}
and therefore
\begin{equation}
    \mathcal{L}_j(\Theta^{+}) > \mathcal{L}_j(\Theta).
\end{equation}
This means that the update induced by view $i$ can increase the reconstruction loss of view $j$. 
Therefore, single-view optimization is not conflict-free. 
Rather, the conflict between view $i$ and view $j$ is unobservable at the current iteration because $g_j$ is not computed. 
Such a conflict can only be implicitly revealed in later iterations when view $j$ is sampled.

\subsection{Detailed Derivation of Harmonized Gradients}
\label{app:harmonized_gradient_derivation}
Assume that $g_1$ and $g_2$ are nonzero and not exactly collinear. 
When $g_1^\top g_2<0$, the angle between them is
\begin{equation}
    \theta
    =
    \arccos
    \left(
    \frac{g_1^\top g_2}{\|g_1\|\|g_2\|}
    \right)
    \in
    \left(\frac{\pi}{2},\pi\right).
\end{equation}
We construct an orthonormal basis $\{u_1,u_2\}$ for the two-dimensional subspace $\mathcal{V}$:
\begin{equation}
    u_1 = \frac{g_1}{\|g_1\|},
    \quad
    u_2 =
    \frac{
    g_2-(g_2^\top u_1)u_1
    }{
    \|g_2-(g_2^\top u_1)u_1\|
    }.
\end{equation}
Since
\begin{equation}
    g_2^\top u_1 = \|g_2\|\cos\theta,
\end{equation}
the original gradients can be represented as
\begin{equation}
    g_1 = \|g_1\|u_1,
    \quad
    g_2 =
    \|g_2\|
    \left(
    \cos\theta\,u_1+\sin\theta\,u_2
    \right).
\end{equation}

The original angular separation between $g_1$ and $g_2$ is $\theta>\frac{\pi}{2}$. 
To remove the negative inner-product conflict, their final angular separation should be $\frac{\pi}{2}$. 
Thus, the required angular correction is
\begin{equation}
    \Delta\theta = \theta-\frac{\pi}{2}.
\end{equation}
We introduce an allocation parameter $\rho\in[0,1]$ to distribute this correction between the two gradients:
\begin{equation}
    \beta = \rho\left(\theta-\frac{\pi}{2}\right),
    \quad
    \bar{\beta}
    =
    (1-\rho)\left(\theta-\frac{\pi}{2}\right).
\end{equation}
Here, $g_1$ is rotated from angle $0$ to angle $\beta$ in the basis $\{u_1,u_2\}$, while $g_2$ is rotated from angle $\theta$ toward $g_1$ by $\bar{\beta}$. 
Therefore, the final angle of $\tilde{g}_2$ with respect to $u_1$ is
\begin{equation}
    \theta-\bar{\beta}
    =
    \theta
    -
    (1-\rho)
    \left(\theta-\frac{\pi}{2}\right).
\end{equation}
Since $\beta=\rho(\theta-\frac{\pi}{2})$, we have
\begin{equation}
    \theta-\bar{\beta}
    =
    \frac{\pi}{2}+\beta.
\end{equation}
Therefore, the harmonized gradients are
\begin{equation}
    \tilde{g}_1
    =
    \|g_1\|
    \left(
    \cos\beta\,u_1
    +
    \sin\beta\,u_2
    \right),
\end{equation}
and
\begin{equation}
    \tilde{g}_2
    =
    \|g_2\|
    \left(
    \cos\left(\frac{\pi}{2}+\beta\right)u_1
    +
    \sin\left(\frac{\pi}{2}+\beta\right)u_2
    \right).
\end{equation}
Using
\begin{equation}
    \cos\left(\frac{\pi}{2}+\beta\right)=-\sin\beta,
    \quad
    \sin\left(\frac{\pi}{2}+\beta\right)=\cos\beta,
\end{equation}
we obtain
\begin{equation}
    \tilde{g}_2
    =
    \|g_2\|
    \left(
    -\sin\beta\,u_1
    +
    \cos\beta\,u_2
    \right).
\end{equation}
This construction preserves the magnitudes of the two original gradients while changing only their directions within $\mathcal{V}$.

\subsection{Orthogonality Verification}
\label{app:orthogonality_verification}
We now verify that the two harmonized gradients are orthogonal. 
Since $\{u_1,u_2\}$ is an orthonormal basis, we have
\begin{equation}
    u_1^\top u_1 = 1,
    \quad
    u_2^\top u_2 = 1,
    \quad
    u_1^\top u_2 = 0.
\end{equation}
Substituting the expressions of $\tilde{g}_1$ and $\tilde{g}_2$, we have
\begin{align}
    \tilde{g}_1^\top \tilde{g}_2
    &=
    \|g_1\|\|g_2\|
    \left(
    \cos\beta\,u_1+\sin\beta\,u_2
    \right)^\top
    \left(
    -\sin\beta\,u_1+\cos\beta\,u_2
    \right) \\
    &=
    \|g_1\|\|g_2\|
    \Big(
    -\cos\beta\sin\beta\,u_1^\top u_1
    +
    \cos^2\beta\,u_1^\top u_2
    -
    \sin^2\beta\,u_2^\top u_1
    +
    \sin\beta\cos\beta\,u_2^\top u_2
    \Big) \\
    &=
    \|g_1\|\|g_2\|
    \Big(
    -\cos\beta\sin\beta
    +
    \sin\beta\cos\beta
    \Big) \\
    &= 0.
\end{align}
Therefore, the harmonized gradients satisfy
\begin{equation}
    \tilde{g}_1^\top \tilde{g}_2 = 0.
\end{equation}
This confirms that the proposed rotation removes the negative inner-product conflict between the sampled gradients.

\subsection{Recombination in the Original Gradient Basis}
\label{app:gradient_basis_recombination}
The final update direction is defined as the sum of the two harmonized gradients:
\begin{equation}
    \tilde{g}
    =
    \tilde{g}_1+\tilde{g}_2.
\end{equation}
Substituting the expressions of $\tilde{g}_1$ and $\tilde{g}_2$, we obtain
\begin{align}
    \tilde{g}
    &=
    \|g_1\|
    \left(
    \cos\beta\,u_1+\sin\beta\,u_2
    \right)
    +
    \|g_2\|
    \left(
    -\sin\beta\,u_1+\cos\beta\,u_2
    \right) \\
    &=
    \left(
    \|g_1\|\cos\beta
    -
    \|g_2\|\sin\beta
    \right)u_1
    +
    \left(
    \|g_1\|\sin\beta
    +
    \|g_2\|\cos\beta
    \right)u_2.
\end{align}
Since both $\tilde{g}_1$ and $\tilde{g}_2$ lie in $\mathrm{span}(g_1,g_2)$, their sum can be expressed as a linear combination of the original gradients:
\begin{equation}
    \tilde{g}
    =
    \tau_1 g_1+\tau_2 g_2.
\end{equation}
Using
\begin{equation}
    g_1 = \|g_1\|u_1,
    \quad
    g_2 =
    \|g_2\|
    \left(
    \cos\theta\,u_1+\sin\theta\,u_2
    \right),
\end{equation}
we have
\begin{align}
    \tau_1 g_1+\tau_2 g_2
    &=
    \left(
    \tau_1\|g_1\|
    +
    \tau_2\|g_2\|\cos\theta
    \right)u_1
    +
    \tau_2\|g_2\|\sin\theta\,u_2.
\end{align}
By matching the coefficient of $u_2$, we obtain
\begin{equation}
    \tau_2\|g_2\|\sin\theta
    =
    \|g_1\|\sin\beta+\|g_2\|\cos\beta,
\end{equation}
which gives
\begin{equation}
    \tau_2
    =
    \frac{\|g_1\|\sin\beta}{\|g_2\|\sin\theta}
    +
    \frac{\cos\beta}{\sin\theta}.
\end{equation}
By matching the coefficient of $u_1$, we have
\begin{equation}
    \tau_1\|g_1\|
    +
    \tau_2\|g_2\|\cos\theta
    =
    \|g_1\|\cos\beta-\|g_2\|\sin\beta.
\end{equation}
Therefore,
\begin{equation}
    \tau_1
    =
    \frac{
    \|g_1\|\cos\beta
    -
    \|g_2\|\sin\beta
    -
    \tau_2\|g_2\|\cos\theta
    }{
    \|g_1\|
    }.
\end{equation}
Substituting the expression of $\tau_2$ and simplifying, we obtain
\begin{equation}
    \tau_1
    =
    \frac{\sin(\theta-\beta)}{\sin\theta}
    -
    \frac{\|g_2\|\cos(\theta-\beta)}{\|g_1\|\sin\theta}.
\end{equation}
Thus, the harmonized update can be implemented as
\begin{equation}
    \tilde{g}
    =
    \tau_1 g_1+\tau_2 g_2,
\end{equation}
where
\begin{equation}
    \tau_1
    =
    \frac{\sin(\theta-\beta)}{\sin\theta}
    -
    \frac{\|g_2\|\cos(\theta-\beta)}{\|g_1\|\sin\theta},
    \quad
    \tau_2
    =
    \frac{\|g_1\|\sin\beta}{\|g_2\|\sin\theta}
    +
    \frac{\cos\beta}{\sin\theta}.
\end{equation}
When no conflict exists, i.e., $g_1^\top g_2\geq 0$, we do not perform harmonization and simply set $\tau_1=\tau_2=1$.

\subsection{Expected Conflict Coverage over the Training Set}
\label{app:expected_conflict_coverage}
We further analyze how random two-view sampling covers pairwise conflicts over the training set. 
Let $\mathcal{P} = \{(i,j)\mid 1\leq i<j\leq N\}$ be the set of all unordered view pairs, and let $M = |\mathcal{P}| = \frac{N(N-1)}{2}$ be the number of all possible view pairs. Assume that one unordered view pair is sampled uniformly at random at each iteration. 
For a fixed pair $(i,j)$, the probability that it is sampled in one iteration is
\begin{equation}
    p = \frac{1}{M}.
\end{equation}
Therefore, the probability that this pair is not sampled in one iteration is
\begin{equation}
    1-p = 1-\frac{1}{M}.
\end{equation}
After $T$ independent training iterations, the probability that this pair has never been sampled is
\begin{equation}
    \left(1-\frac{1}{M}\right)^T.
\end{equation}
Consequently, the probability that this pair has been sampled at least once after $T$ iterations is
\begin{equation}
    P_{\mathrm{cover}}(T)
    =
    1-\left(1-\frac{1}{M}\right)^T.
    \label{eq:pair_coverage}
\end{equation}
When $M$ is large, this can be approximated by
\begin{equation}
    P_{\mathrm{cover}}(T)
    \approx
    1-\exp\left(-\frac{T}{M}\right).
\end{equation}
This probability also gives the expected fraction of all view pairs that have been sampled at least once:
\begin{equation}
    \mathbb{E}
    \left[
    \frac{\#\mathrm{covered\ pairs}}{M}
    \right]
    =
    1-\left(1-\frac{1}{M}\right)^T.
\end{equation}
Therefore, random two-view sampling progressively covers the pairwise view space as training proceeds. To reach a target pairwise coverage ratio $q$, the required number of iterations satisfies
\begin{equation}
    T
    \ge
    \frac{\log(1-q)}
    {\log(1-\frac{1}{M})}.
\end{equation}
For large $M$, this is approximately
\begin{equation}
    T
    \gtrsim
    M\log\frac{1}{1-q}.
\end{equation}
Therefore, reaching $50\%$, $80\%$, $95\%$, $99\%$, and $99.9\%$ expected pair coverage requires approximately
\begin{equation}
T_{50}\approx 0.7M, \quad
T_{80}\approx 1.6M, \quad
T_{95}\approx 3M,\quad
T_{99}\approx 4.6M,\quad
T_{99.9}\approx 6.9M.
\end{equation}
At this stage, most pairwise relationships have been exposed at least once in expectation.

\section{Broader Impacts}
\label{app:broader_impacts}
This work aims to improve the robustness of 3D Gaussian Splatting for in-the-wild scene reconstruction. By reducing the influence of transient distractors and cross-view inconsistencies, the proposed method can benefit real-world applications that require reliable novel view synthesis, such as virtual reality, digital content creation, cultural heritage preservation, and robotic perception in unconstrained environments.

At the same time, more robust scene reconstruction techniques may also raise potential concerns. For example, reconstructing scenes from unconstrained image collections could involve privacy-sensitive environments or unintentionally capture people, vehicles, or private properties. Such risks are not specific to our method, but are inherent to scene reconstruction technologies when applied to real-world data. We encourage responsible data collection and deployment practices, including obtaining proper consent, avoiding sensitive scenarios, and following applicable privacy regulations.


\end{document}